\documentclass[pdflatex,sn-nature]{sn-jnl}

\usepackage{graphicx}%
\usepackage{multirow}%
\usepackage{amsmath,amssymb,amsfonts}%
\usepackage{amsthm}%
\usepackage{mathrsfs}%
\usepackage[title]{appendix}%
\usepackage{xcolor}%
\usepackage{textcomp}%
\usepackage{manyfoot}%
\usepackage{booktabs}%
\usepackage{algorithm}%
\usepackage{algorithmicx}%
\usepackage{algpseudocode}%
\usepackage{listings}%
\usepackage{gensymb}
\usepackage{comment}
\usepackage{makecell}
\usepackage{threeparttable}

\theoremstyle{thmstyleone}%
\theoremstyle{thmstyletwo}%

\theoremstyle{thmstylethree}%

\begin{document}

\title[Article Title]{Reliable End-to-End Material Information Extraction from the Literature with Source-Tracked Multi-Stage Large Language Models}



\author[1]{\fnm{Xin} \sur{Wang}}\email{xwang218@crimson.ua.edu}

\author[2]{\fnm{Anshu} \sur{Raj}}\email{Anshu.Raj-1@ou.edu}

\author[3]{\fnm{Matthew} \sur{Luebbe}}\email{mjlvp8@mst.edu}

\author[3]{\fnm{Haiming} \sur{Wen}}\email{wenha@mst.edu}

\author[2]{\fnm{Shuozhi} \sur{Xu}}\email{shuozhixu@ou.edu}

\author[1]{\fnm{Kun} \sur{Lu}}\email{klu@ua.edu}

\affil[1]{\orgdiv{School of Library and Information Studies}, \orgname{University of Alabama}, \orgaddress{\city{Tuscaloosa}, \postcode{35487}, \state{AL}, \country{USA}}}

\affil[2]{\orgdiv{School of Aerospace and Mechanical Engineering}, \orgname{University of Oklahoma},  \orgaddress{\city{Norman}, \postcode{73019}, \state{OK}, \country{USA}}}

\affil[3]{\orgdiv{Department of Materials Science and Engineering}, \orgname{Missouri University of Science and Technology}, \orgaddress{\city{Rolla}, \postcode{65409}, \state{MO}, \country{USA}}}


\abstract{Data-driven materials discovery requires large-scale experimental datasets, yet most of the information remains trapped in unstructured literature. Existing extraction efforts often focus on a limited set of features and have not addressed the integrated composition–processing–microstructure–property relationships essential for understanding materials behavior, thereby posing challenges for building comprehensive databases. To address this gap, we propose a multi-stage information extraction pipeline powered by large language models, which captures 47 features spanning composition, processing, microstructure, and properties exclusively from experimentally reported materials. The pipeline integrates iterative extraction with source tracking to enhance both accuracy and reliability. Evaluations at the feature level (independent attributes) and tuple level (interdependent features) yielded F1 scores around 0.96. Compared with single-pass extraction without source tracking, our approach improved F1 scores of microstructure category by 10.0\% (feature level) and 13.7\% (tuple level), and reduced missed materials from 49 to 13 out of 396 materials in 100 articles on precipitate-containing multi-principal element alloys (miss rate reduced from 12.4\% to 3.3\%). The pipeline enables scalable and efficient literature mining, producing databases with high precision, minimal omissions, and zero false positives. These datasets provide trustworthy inputs for machine learning and materials informatics, while the modular design generalizes to diverse material classes, enabling comprehensive materials information extraction.}




\maketitle

\section{Introduction}\label{sec1}

Materials science relies on a comprehensive understanding of processing, microstructure, property, and performance, which together form the well-established materials tetrahedron \cite{de2019new, deagen2022materials, jain2013commentary}. Extracting such information from the rapidly growing body of literature is critical to accelerate data-driven discovery. Computational databases such as Materials Project ($>$200,000 materials), MaterialsAtlas.org ($>$130,000 compounds), AFLOW ($>$3.5 million entries), and NOMAD ($>$100 million calculations) have structured vast amounts of simulation results \cite{hu2022materialsatlas, sbailo2022nomad, curtarolo2012aflow}. In contrast, while some experimental databases exist (such as HTEM\ \cite{zakutayev2018open}, RRUFF\ \cite{lafuente2015},  SpringerMaterials), the vast majority of experimental information, including processing histories, microstructures, and properties, remains largely trapped in unstructured text across millions of publications. Recent analyses reveal that materials science publications grow at 4.10\% annually and were doubled within 17.3 years \cite{bornmann2021growth}, creating an ever-expanding yet underutilized repository of knowledge.

Within this broader challenge, microstructure information is critical but difficult to extract. Microstructure directly governs macroscopic behavior, from mechanical strength and ductility to thermal and electrical performance \cite{schmidt2019recent, groeber2014dream}. Previous studies have attempted to characterize microstructure using image analysis \cite{zhao2023new, decost2015computer, paudel2024priming, calvat2025learning}, but such approaches cannot capture the contextual information available in text, where microstructural details are often reported alongside composition, processing, and properties. A complete description involves not only chemical composition and processing parameters, but also phase structure, precipitate morphology, and property outcomes under diverse conditions \cite{zhu2024microstructure, kalinin2015big}. Unlike other categories of material information, microstructural details are highly multidimensional, scattered throughout text, and often referenced indirectly through abbreviations or symbols, making it a bottleneck in tracking the complete processing–microstructure–property chain. This gap highlights why comprehensive information extraction cannot be achieved without microstructure. Without comprehensive information extraction, it will be difficult to realize data-driven discovery and accelerated innovation in materials science \cite{wodo2016microstructural}.

Traditional natural language processing (NLP) approaches have made significant contributions to materials information extraction. Rule-based systems such as OSCAR4 \cite{jessop2011oscar4}, ChemicalTagger \cite{hawizy2011chemicaltagger}, and ChemDataExtractor \cite{swain2016chemdataextractor} specialized in automated extraction of chemical entities and relationships from scientific literature, establishing important foundations for the field. These tools have been widely adopted and alloy-focused extensions have demonstrated the feasibility of domain-specific extraction \cite{pfeiffer2022aluminum}. Building on these advances, ChemDataExtractor2 expanded capabilities to include quantitative property extraction \cite{mavracic2021chemdataextractor}.

In parallel, machine learning approaches introduced semi-supervised and shallow neural models to tackle extraction challenges. Huo et al. \cite{huo2019semi} applied topic modeling and Markov chains for synthesis procedure classification, while Yan et al. \cite{yan2022materials} employed weak supervision frameworks such as Snorkel to reduce annotation requirements.

These foundational works have paved the way for more comprehensive extraction frameworks, though limitations remain. Rule-based systems struggle with the complexity and variability of scientific language. Machine learning approaches, while more flexible, require extensive manual annotation and continue to struggle with domain-specific terminology and complex multi-dimensional relationships inherent in materials science literature. 

A major breakthrough came with the use of distributed representations. Methods such as word2vec \cite{mikolov2013distributed}, doc2vec \cite{le2014distributed}, and BERT (bidirectional encoder representations from transformers) \cite{devlin2019bert} enabled models to capture latent semantic relationships in materials text. This paradigm demonstrated success in unsupervised discovery of novel materials from millions of abstracts \cite{tshitoyan2019unsupervised}. Later work has extended these approaches to full-text analysis. MatSciBERT achieved state-of-the-art results on materials science NER (named entity recognition) and relation classification tasks \cite{gupta2022matscibert}. MaterialBERT demonstrated effective word embeddings for materials clustering and functional group relationships\ \cite{yoshitake2022materialbert}. And MaterialsBERT developed a comprehensive pipeline for extracting material property records from polymer literature\ \cite{shetty2023general}. Nevertheless, these models are fundamentally representational, requiring downstream fine-tuning, task-specific annotations, or handcrafted rules to convert embeddings into structured knowledge.

Generative large language models (LLMs) such as GPT (generative pre-trained transformer) differ from the previous models in that they can directly produce structured outputs and capture relationships beyond the sentence level without fine-tuning. Recent studies have demonstrated their potential for automated information extraction in materials science. Zheng et al.\ \cite{zheng2023chatgpt} used a ChatGPT-based chemistry assistant to mine synthesis parameters of metal–organic frameworks. Gupta et al.\ \cite{gupta2024data} combined LLMs with classifiers to curate large polymer property databases. Torres da Silva et al.\ \cite{da2024automated} structured synthesis protocols of Metal–Organic Frameworks, Covalent Organic Frameworks, and zeolites directly from full texts (in PDF format), while Thway et al.\ \cite{thway2024harnessing} parsed solid-state synthesis ``recipes'' from text. Other studies have emphasized reliability: Dagdelen et al.\ \cite{dagdelen2024structured} fine-tuned LLMs for joint entity–relation extraction, and Polak and Morgan \cite{polak2024extracting} developed ChatExtract to reduce hallucinations and improve precision. Wang et al.\ \cite{wang2024does} found that GPT-4 substantially outperformed traditional methods in band gap extraction tasks, and with prompt engineering, its accuracy could be improved to 93.98\%. Yet, existing work remains limited to single dimensions, simple composition–property relations, or simple composition-processing correspondence, restricting the understanding of the full composition–processing–microstructure–property relationships.

\begin{figure}[h]
    \centering
\includegraphics[width=0.9\textwidth, height=0.7\textheight]{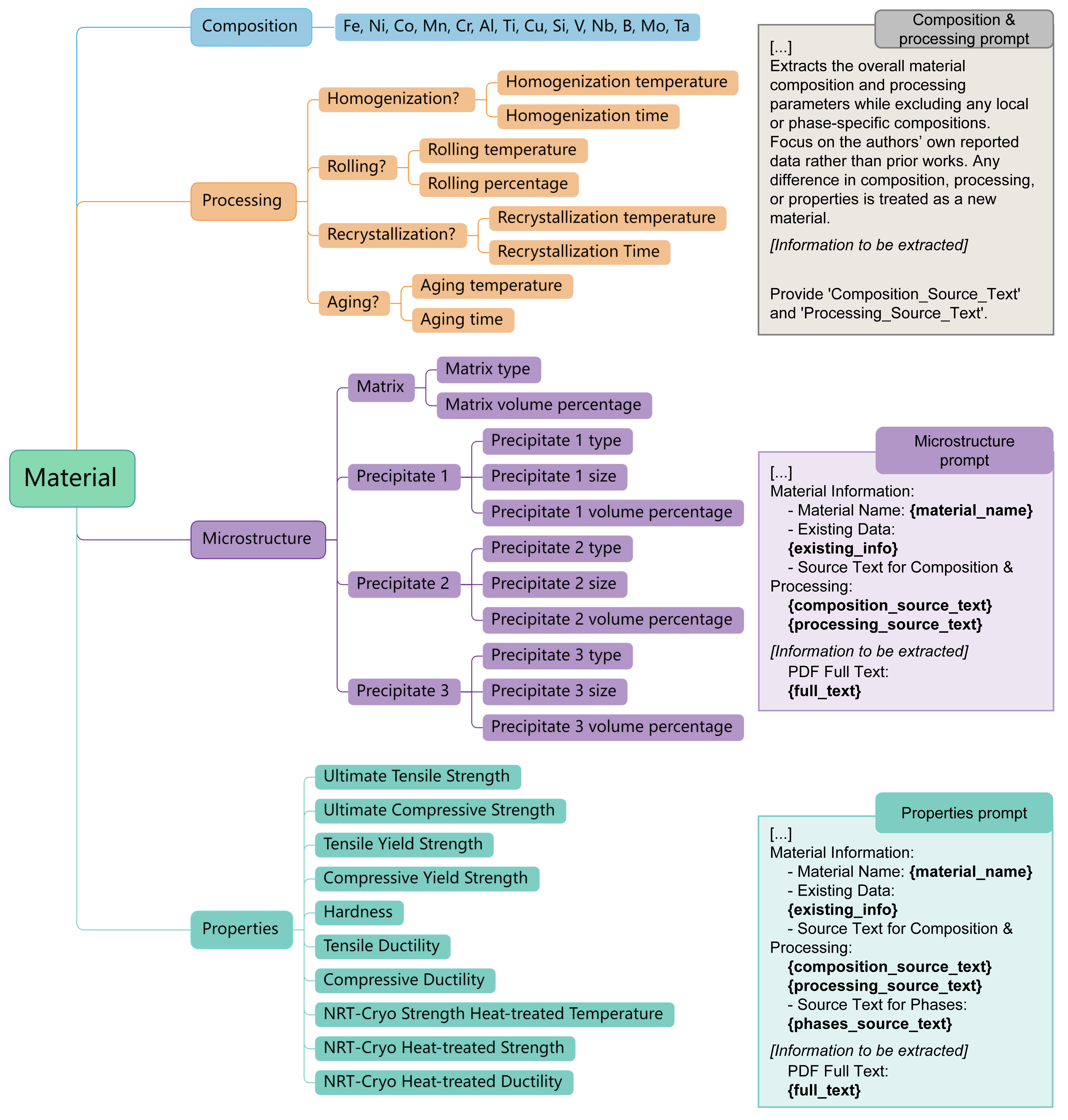}
    \caption{Extraction framework showing the 47 features extracted across four categories (composition, processing, microstructure, and properties) and the specialized prompt design for each category. Here, ``NRT-Cryo'' means non-room temperature cryogenic conditions.}
    \label{fig:extracted_information_2}
\end{figure}

This study proposes a multi-stage extraction pipeline to capture the hierarchical nature of materials information across the complete composition-processing-microstructure-property chain. While the well-established materials tetrahedron is typically defined by processing, structure, properties, and performance, we instead emphasize composition-processing-microstructure-property relationships. Performance often depends on system-level factors beyond intrinsic materials descriptors, whereas composition plays a foundational role in governing processing routes and determining structure–property relations, making it more suitable for data-driven discovery. We target 47 distinct features spanning these four categories: 14 composition features, 12 processing parameters, 11 microstructure attributes, and 10 material properties. Figure~\ref{fig:extracted_information_2} shows what features are extracted and how prompts are designed for each category. In this paper, our framework is applied to 100 journal articles on the topic of ``precipitate-containing multi-principal element alloys,'' because these materials contain rich information on both composition and microstructure. These 100 articles contain original experimental results. While some papers also include modeling or secondary analyses, we instructed GPT to extract only experimentally reported data.

\begin{figure}[h]
    \centering
    \includegraphics[width=1.0\textwidth]{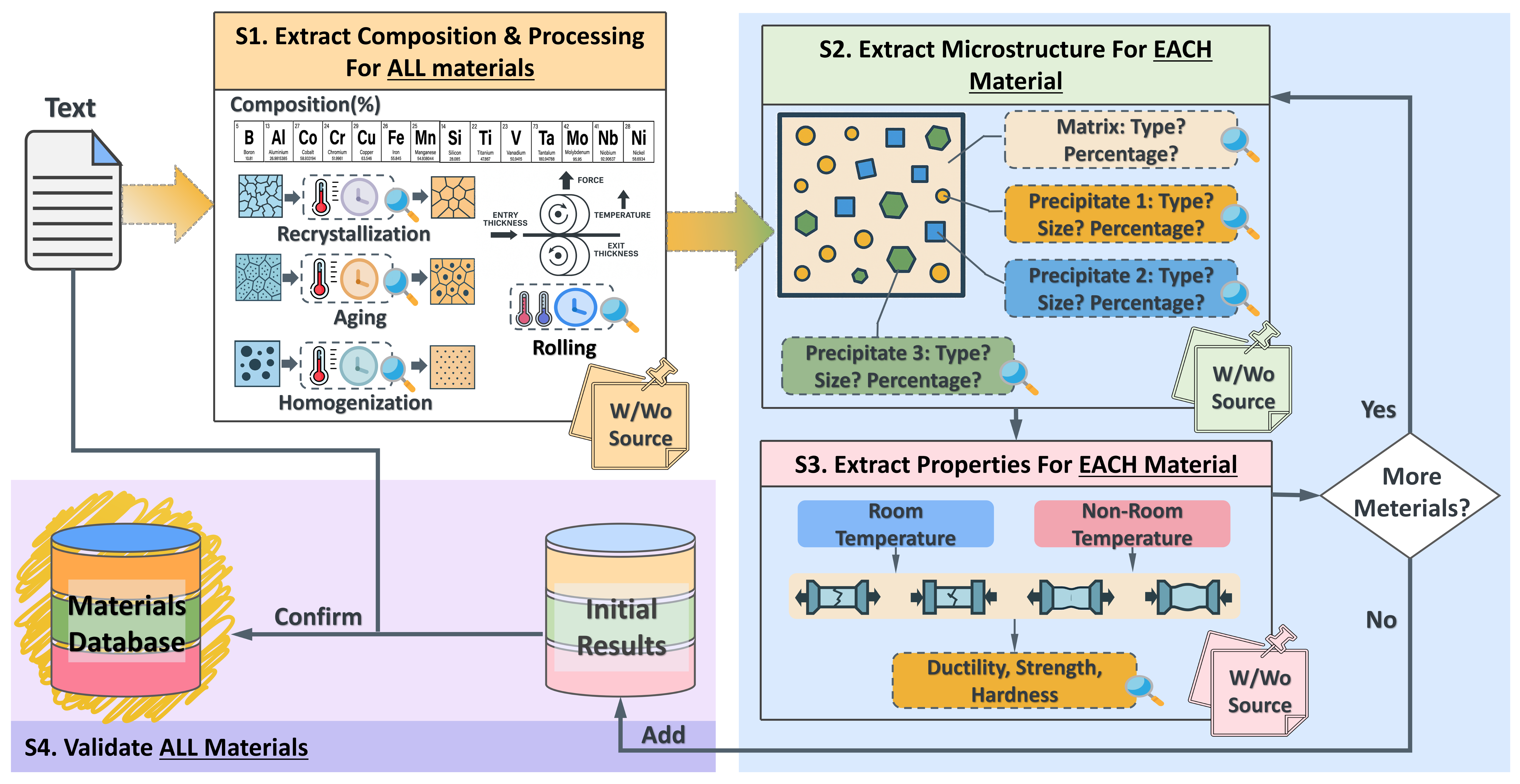}
    \caption{Architecture of our multi-stage extraction pipelines (with/without source tracking).}
    \label{fig:pipeline-architecture}
\end{figure}

As shown in Figure~\ref{fig:pipeline-architecture}, our pipeline employs a four-stage, large language model (LLM)–driven approach. Stage 1, performed by the LLM, extracts all materials along with their chemical compositions and processing parameters from the entire article text. All extracted materials are derived directly from the authors’ own experimental results, ensuring that the database reflects only primary research findings. Following this initial extraction, the LLM iteratively processes each identified material: Stage 2 extracts microstructure information, including matrix and precipitate types, sizes, and volume fractions, using the composition and processing context from Stage 1. Stage 3 then extracts mechanical properties for the same material while leveraging all previously accumulated information. This Stage 2–3 sequence repeats for each material identified in Stage 1. Only after completing the extraction for all materials does Stage 4, again executed by the LLM, perform comprehensive validation using the complete material database and original source texts. This database contains all materials identified in Stage 1 along with their extracted composition, processing, phase structure, and property information from Stages 2–3.

A key innovation of our approach is the source tracking mechanism, which requires LLMs to provide an explicit reasoning basis for each extracted value in every stage. This mechanism preserves links between extracted values and their original text sources, reducing reference loss during multi-stage processing and enabling systematic error correction through traceable validation. The source tracking mechanism requires multi-stage processing to maintain contextual links across different information categories, distinguishing it from single-pass approaches. Thus, to demonstrate the effectiveness of this mechanism and the necessity of multi-stage processing, we developed three pipeline variants: single-pass extraction without source tracking, multi-stage extraction without source tracking, and multi-stage extraction with source tracking. 

These pipelines are evaluated by two complementary approaches: feature-level evaluation that treats each feature independently, while tuple-level evaluation that groups interdependent features and evaluates them holistically. Using OpenAI's o3-mini model with prompt engineering and confirmation strategies, our approach achieves a 0.959 F1 score in feature-level evaluation. Tuple-level evaluation demonstrates nearly identical performance levels, confirming the robustness of our extraction methodology. Even in the most challenging microstructure extraction task, our pipeline maintains strong performance with an F1 Score of 0.950 and an accuracy of 0.966 in feature-level evaluation, and an F1 Score of 0.924 and an accuracy of 0.923 in tuple-level evaluation. Our work shows consistent accuracy across all feature categories critical for materials database construction.

\section{Results}\label{sec2}

\subsection{Overall Performance Evaluation}

\begin{figure}[h]
    \centering
    \includegraphics[width=1\textwidth]{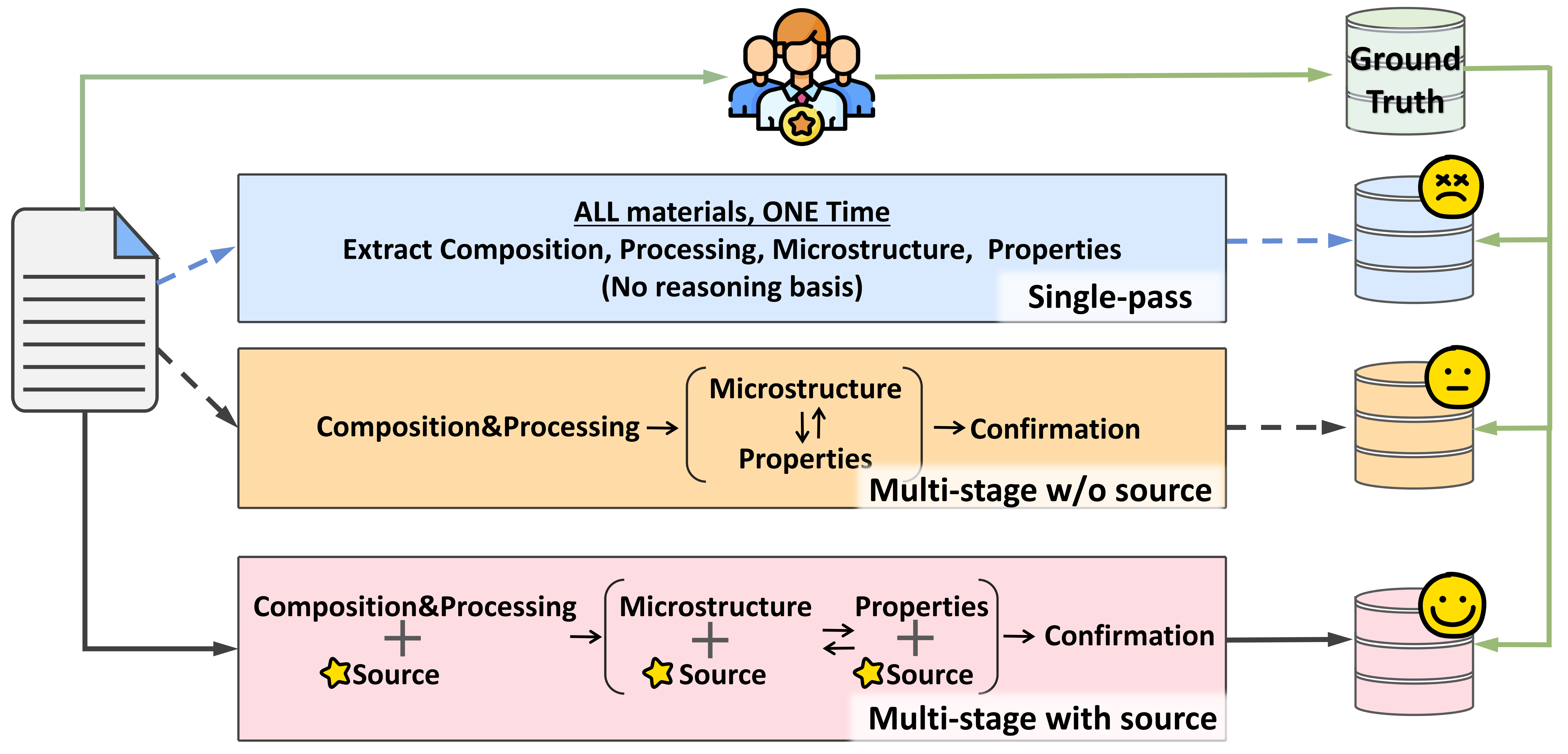}
    \caption{Processes of three extraction pipelines.}
    \label{fig:pipeline-comparison}
\end{figure}

Three extraction pipelines are compared to understand the impact of the multi-stage pipeline and source tracking in material information extraction. Figure~\ref{fig:pipeline-comparison} illustrates their processes: (1) single-pass extraction without source tracking, (2) multi-stage extraction without source tracking, and (3) multi-stage extraction with source tracking. The fundamental distinction is in how information flows through the system and whether contextual source text is preserved throughout the extraction process.

To evaluate the performance, the ground truth was developed through a multi-expert validation process involving five co-authors. Initially, two domain experts manually extracted material information from the literature to create a preliminary version of the ground truth. To account for human errors, a third co-author with expertise in information extraction reviewed and marked discrepancies between the preliminary ground truth and the LLM-generated output. A fourth domain expert then conducted a review by comparing both versions against the original literature, identifying any missed extractions, erroneous additions, or inconsistencies. The reconciled results were subsequently checked again by the third co-author. Finally, the fourth and fifth experts finalized the ground truth dataset. This multi-stage process ensured the rigor and reliability of the ground truth.

Building on this multi-expert process, we further established rules for ground truth representation and feature value assignment. Our ground truth defines positive cases as values that are explicitly present in the original text or correct values obtained through valid reasoning, while negative cases refer to values that are absent or not mentioned in the text. The confusion matrix classification follows standard definitions:
\begin{itemize}
    \item True Positives (TP): ground truth and extracted values are present and match.
    \item True Negatives (TN): both ground truth and extracted values are absent.
    \item False Negatives (FN): ground truth has a value, but the extraction is empty.
    \item False Positives (FP): either the ground truth is missing while the extraction provides a value, or both contain values, but the extracted one is incorrect.
\end{itemize}

In feature value assignment, all elements not reported in the composition are assigned a value of zero. For processing features, we implemented a specific approach to handle missing information: when papers do not mention some processing steps (homogenization, aging, or recrystallization), we set the corresponding boolean indicator to 0 (e.g., \texttt{Aging? = 0}) and assign null values to associated temperature and time. For microstructure features, any precipitate or matrix information not explicitly mentioned in the text is marked as ``not reported'', including their corresponding volume fractions and sizes. Similarly, material properties not reported are also marked as ``not reported''.

Table~\ref{tab:performance_comparison} presents a comparison of performance metrics across three pipelines using feature-level and tuple-level evaluation. Performance metrics are calculated by treating each feature independently and then aggregating the results across all features and materials. The evaluation penalizes the system for any missed materials (counting all 47 features as false negatives for that material) or erroneously extra materials (counting all 47 features as false positives for that material).

\begin{table}[htbp]
\centering
\footnotesize
\caption{Feature-level and tuple-level evaluation performance comparison of three extraction pipelines (With Extra/Missed). Bold values indicate the best performance for each metric.}
\label{tab:performance_comparison}
\begin{tabular}{l|ccc|ccc}
\toprule
 & \multicolumn{3}{c}{Feature-level} & \multicolumn{3}{c}{Tuple-level} \\
\cmidrule(lr){2-4} \cmidrule(lr){5-7}
 & Single-pass & \makecell{Multi-stage \\ without source} & \makecell{Multi-stage \\ with source} 
 & Single-pass & \makecell{Multi-stage \\ without source} & \makecell{Multi-stage \\ with source} \\
\midrule
Precision 
 & 0.914 & 0.972 & \textbf{0.978} 
 & 0.920 & 0.971 & \textbf{0.977} \\
Recall 
 & 0.828 & 0.814 & \textbf{0.940} 
 & 0.843 & 0.830 & \textbf{0.947} \\
F1 Score 
 & 0.869 & 0.886 & \textbf{0.959} 
 & 0.880 & 0.895 & \textbf{0.962} \\
Accuracy 
 & 0.845 & 0.864 & \textbf{0.950} 
 & 0.842 & 0.860 & \textbf{0.948} \\
\bottomrule
\end{tabular}
\end{table}

Our results demonstrate that the multi-stage pipeline with source tracking achieves the best performance across all metrics in both evaluation approaches. It achieves F1 scores of 0.959 (feature-level) and 0.962 (tuple-level), while the single-pass pipeline shows F1 scores of 0.869 (feature-level) and 0.880 (tuple-level). Compared to our best approach (multi-stage with source tracking), the single-pass pipeline shows performance gaps of 9.0 percentage points in feature-level evaluation and 8.2 percentage points in tuple-level evaluation.

The primary advantage of multi-stage pipeline with source tracking lies in recall performance, where our best approach achieves 0.940 and 0.947 recall in feature-level and tuple-based evaluation, respectively. The other approaches achieve reacalls at the range of [0.814, 0.843]. The multi-stage processing without source tracking achieved improvements of 1.5 and 1.7\% in F1 score but actually decreased recall performance by 1.3--1.4\% compared to single-pass extraction. However, once source tracking is incorporated, both F1 score and recall improvements become substantial, with F1 score improving by 6.7--7.3\% and recall improving by 11.7--12.6\%. These huge gaps demonstrate the critical importance of maintaining source context throughout the extraction pipeline.

\subsection{Material Extraction Completeness Analysis}

The ability to correctly identify and extract all materials is essential for building comprehensive databases and enabling data-driven materials discovery. 
Among the 396 materials present in our testing corpus, the source-tracking pipeline missed only 13 materials (3.3\%), while the multi-stage pipeline without source tracking missed 43 materials (10.8\%), and the single-pass extraction both missed 37 materials and incorrectly identified 12 non-existent materials, resulting in 49 total errors (12.4\%).


The distribution of missed materials offers insights into the failure modes of each approach. For the multi-stage pipeline with source tracking, the 13 missed materials predominantly fall into two categories:

\begin{itemize}
    \item \textbf{Base material overlooked in favor of processed variants (7 cases):} These involve as-manufactured materials that serve as the starting point for subsequent processing but lack explicit processing procedure descriptors. The extraction system prioritized materials with clear processing parameters (e.g., ``aged at 900\textdegree C for 50h'') while treating unprocessed base materials as mere reference states rather than independent research objects.
    
    \item \textbf{Materials with implicit processing parameters (6 cases):} The system missed materials where processing parameters were implicitly expressed rather than explicitly identified as separate research objects. For example, the text primarily focused on the 2-hour annealed material (``HGS-HP MEA annealed for 2 hours'') as the main research object, while the 30-minute condition was mentioned only briefly in the context of microstructural evolution analysis. The extraction system failed to recognize the 30-minute annealed material as a separate, extractable material due to its implicit presentation, resulting in a missed material that was actually studied experimentally.
\end{itemize}

The multi-stage pipeline without source tracking resulted in significant omissions when extracting materials. For instance, when the text stated ``Unlike the FSP condition, the FSA specimen showed different deformation mechanisms,'' the framework without source tracking could not link it to proper material identity. The term ``the FSA specimen'' lost its specific referential meaning without the contextual framework. This led to the complete omission of the FSP condition from the extracted data. Without source anchoring, ambiguous references and pronouns lost their contextual meaning during validation. As a result, the system conflated materials or eliminated them entirely.

In addition to the limitations mentioned above, the single-pass extraction pipeline also suffers from the drawback of producing redundant information. The 12 non-existent materials were extracted due to misunderstandings of historical references or discussions of hypothetical scenarios rather than actual experimental material. When a paper discussed ``Referring to S.\ Singh's investigation, the formation of Al-Ni-rich phase ...'', the single-pass system interpreted this as the authors' own work.

The source-tracking approach, with its 3.3\% miss rate and zero false positive materials, ensures a level of data quality suitable for applications in material informatics. This is particularly important because the extracted materials data are often used for downstream tasks such as building machine learning models to predict material properties. Omission of a substantial number of materials reduces the amount of available training data, while the inclusion of hallucinated materials introduces noise that can lead to model errors. Prior studies have shown that even 10\% label noise can significantly increase the uncertainty and dispersion of predictions, and when noise levels reach several tens of percent, model performance deteriorates markedly~\cite{song2022learning,ding2022impact}. By keeping omissions and hallucinations low, our approach tackles these challenges and yields data of sufficient quality for material informatics applications.

\subsection{Feature Performance Analysis}

The feature performance analysis focuses on the variations of extraction performance across 47 features in four categories. In this analysis, we excluded cases involving missed or extra materials, as each feature extracted from a missed material is directly treated as a false negative (FN), and each feature from an extra material is treated as a false positive (FP). Therefore, we focused this evaluation on correctly identified materials to ensure an isolated assessment of feature extraction quality.

To assess the average performance, we adopted two complementary evaluation strategies. The first approach, called feature-level evaluation, involved calculating a score for each individual feature, such as matrix phase type or ultimate tensile strength, and then averaging these scores within each of the four categories. This evaluation method treats each feature independently and provides an overall view of the model's performance across different features.

The second approach focused on groupings of interdependent features. This tuple-level evaluation was applied specifically to the processing and microstructure categories, where features exhibit strong interdependencies. For processing, we used tuples such as (aging applied?, time, temperature) or (rolling applied?, thickness reduction, temperature). For microstructure, tuples included (precipitate type, size, volume fraction) or (matrix type, volume fraction). A predicted tuple was counted as a true positive (TP) if it matched a ground truth tuple exactly. If a tuple was absent from both predictions and ground truth, it was a true negative (TN). A tuple was a false positive (FP) if any component was incorrect, and a false negative (FN) if it lacked one or more components that were present in the ground truth. Notably, composition and properties features were evaluated only using the single-feature approach, as these features are typically reported and interpreted independently. Consequently, the tuple-level and feature-level evaluation results are the same for composition and properties categories.

\begin{figure}[h]
    \centering
    \includegraphics[width=1\textwidth]{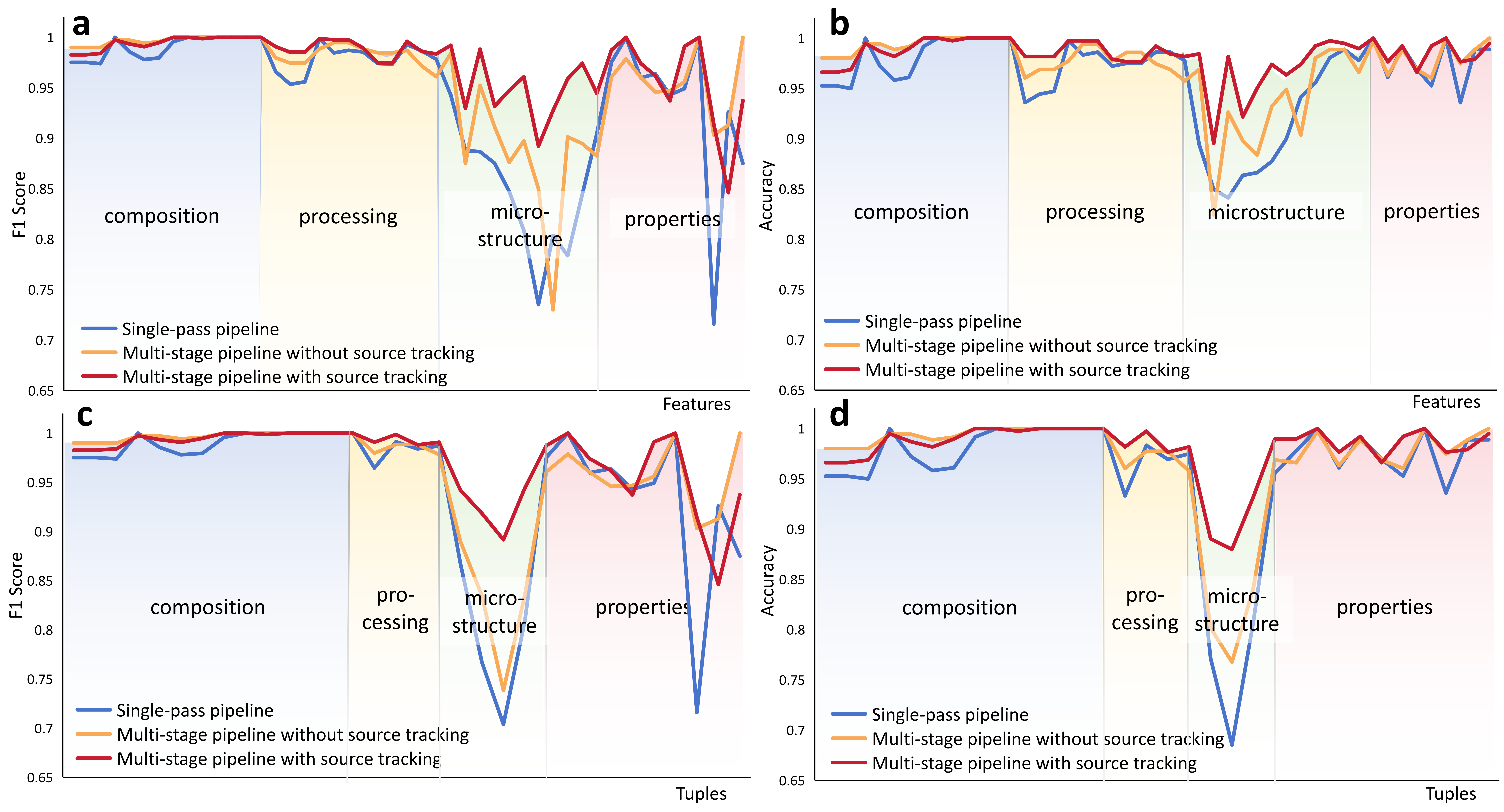}
    \caption{Performance comparison of three extraction pipelines across 47 features organized into composition, processing, microstructure, and properties categories. \textbf{a} Feature-level F1 score comparison showing individual feature extraction accuracy for each pipeline across the four categories. \textbf{b} Feature-level accuracy comparison demonstrating the proportion of correctly extracted features within each category. \textbf{c} Tuple-level F1 score comparison evaluating the extraction of interdependent feature groups, applied specifically to processing and microstructure categories where features exhibit strong relationships. \textbf{d} Tuple-level accuracy comparison measuring the correct extraction of complete feature tuples.}
    \label{fig:feature-metrics}
\end{figure}

Figure \ref{fig:feature-metrics} presents a comparison of the performance of all 47 features under two evaluation strategies. The results reveal performance patterns across four feature categories. In feature-level evaluation (panels a and b), the multi-stage pipeline with source tracking (red line) consistently maintains high performance across all categories, with most features achieving F1 scores and accuracy above 0.95. The multi-stage pipeline without source tracking (orange line) shows moderate performance with some fluctuations, while the single-pass pipeline (blue line) performs noticeably the worst, especially in the microstructure and property categories, where several features score below 0.75. Tuple-level evaluation (panels c and d) exhibits similar trends, but with slightly lower absolute scores, reflecting the increased difficulty of correctly extracting interdependent feature groups. The source-tracking pipeline shows a significant F1 difference only for two non-room-temperature attribute features; for all other features, it achieves either the best performance or performance very close to the best. Overall, the source-tracking pipeline demonstrates strong effectiveness in maintaining contextual relationships among related features. While F1 scores and accuracy provide a balanced perspective on overall performance, detailed precision and recall analysis at the feature level can offer deeper insights into extraction bottlenecks and identify specific failure modes within each category. The following subsections examine precision and recall patterns to better understand the underlying mechanisms driving these performance differences.

\subsubsection{Composition and Processing Features}

All three pipelines showed high recall. For composition features, all three pipelines demonstrated high average recall and precision, as shown in Table \ref{tab:composition_results}. The single-pass pipeline achieved a precision of 0.988 and a recall of 0.993, the multi-stage pipeline without source tracking reached a precision of 0.998 and a recall of 0.996, while the multi-stage pipeline with source tracking attained a precision of 0.992 and a recall of 0.998. 

\begin{table}[htbp]
\centering
\small
\resizebox{\textwidth}{!}{
\begin{tabular}{l|ccc|ccc}
\toprule
 & \multicolumn{3}{c}{Precision} & \multicolumn{3}{c}{Recall} \\
\cmidrule(lr){2-4} \cmidrule(lr){5-7}
 & Single-pass & \makecell{Multi-stage \\ without source} & \makecell{Multi-stage \\ with source} 
 & Single-pass & \makecell{Multi-stage \\ without source} & \makecell{Multi-stage \\ with source} \\
\midrule
Fe  & 0.974 & \textbf{0.994} & 0.974 & 0.977 & 0.986 & \textbf{0.992} \\
Ni  & 0.974 & \textbf{0.994} & 0.974 & 0.977 & 0.986 & \textbf{0.992} \\
Co  & 0.972 & \textbf{0.994} & 0.976 & 0.977 & 0.986 & \textbf{0.992} \\
Mn  & \textbf{1.000} & 0.994 & 0.997 & \textbf{1.000} & \textbf{1.000} & 0.997 \\
Cr  & 0.972 & 0.994 & 0.995 & \textbf{1.000} & \textbf{1.000} & 0.992 \\
Al  & 0.969 & \textbf{1.000} & 0.982 & 0.989 & 0.989 & \textbf{1.000} \\
Ti  & 0.969 & \textbf{1.000} & 0.990 & 0.991 & 0.991 & \textbf{1.000} \\
Cu  & \textbf{1.000} & \textbf{1.000} & \textbf{1.000} & 0.992 & \textbf{1.000} & \textbf{1.000} \\
Si  & \textbf{1.000} & \textbf{1.000} & \textbf{1.000} & \textbf{1.000} & \textbf{1.000} & \textbf{1.000} \\
V   & \textbf{1.000} & \textbf{1.000} & 0.997 & \textbf{1.000} & \textbf{1.000} & \textbf{1.000} \\
Nb  & \textbf{1.000} & \textbf{1.000} & \textbf{1.000} & \textbf{1.000} & \textbf{1.000} & \textbf{1.000} \\
B   & \textbf{1.000} & \textbf{1.000} & \textbf{1.000} & \textbf{1.000} & \textbf{1.000} & \textbf{1.000} \\
Mo  & \textbf{1.000} & \textbf{1.000} & \textbf{1.000} & \textbf{1.000} & \textbf{1.000} & \textbf{1.000} \\
Ta  & \textbf{1.000} & \textbf{1.000} & \textbf{1.000} & \textbf{1.000} & \textbf{1.000} & \textbf{1.000} \\
\midrule
\textbf{Average} 
 & 0.988 & \textbf{0.998} & 0.992 
 & 0.993 & 0.996 & \textbf{0.998} \\
\bottomrule
\end{tabular}
}
\caption{Precision and recall of different pipelines for composition.}
\label{tab:composition_results}
\end{table}


This consistently high performance across all pipelines can be attributed to several factors. First, elemental composition is reported using highly standardized notation and terminology. Second, the extraction task operates within a constrained domain: the periodic table contains just over a hundred elements, which significantly reduces ambiguity. Third, composition data typically appears in well-defined sections of papers, often presented in tables or clearly delineated text passages with explicit quantitative values (e.g., Cr\textsubscript{30}Co\textsubscript{30}Ni\textsubscript{30}Al\textsubscript{5}Ti\textsubscript{5} (at.\%)). In this scenario, even the simplest single-pass approach can achieve near-perfect performance.

Meanwhile, three pipelines demonstrated high recall and precision in extracting processing features under both evaluation methods, with results ranging from 0.968 to 0.995 (Table \ref{tab:process_precision_results} and Table \ref{tab:process_recall_results}). Among them, the multi-stage pipeline with source tracking performed best, achieving a precision of 0.986 and a recall of 0.991 in the feature-level evaluation, and a precision of 0.989 and a recall of 0.995 in the tuple-level evaluation. 


\begin{table}[htbp]
\centering
\small 
\resizebox{\textwidth}{!}{
\begin{tabular}{l|ccc|ccc}
\toprule
 & \multicolumn{3}{c}{Feature level} & \multicolumn{3}{c}{Tuple level} \\
\cmidrule(lr){2-4} \cmidrule(lr){5-7}
 & Single-pass & \makecell{Multi-stage \\ without source} & \makecell{Multi-stage \\ with source} 
 & Single-pass & \makecell{Multi-stage \\ without source} & \makecell{Multi-stage \\ with source} \\
\midrule

\multicolumn{7}{l}{\textbf{Homogenization}} \\
Homogenization?   & 0.934 & 0.977 & \textbf{0.982} & \multirow{3}{*}{0.931} & \multirow{3}{*}{0.977} & \multirow{3}{*}{\textbf{0.982}} \\
Homogenization temperature        & 0.913 & 0.964 & \textbf{0.972} &  &  &  \\
Homogenization time        & 0.917 & 0.964 & \textbf{0.972} &  &  &  \\
\cmidrule(lr){1-7}

\multicolumn{7}{l}{\textbf{Rolling}} \\
Rolling?    & 0.997 & 0.994 & \textbf{0.997} & \multirow{3}{*}{0.994} & \multirow{3}{*}{0.994} & \multirow{3}{*}{\textbf{0.997}} \\
Rolling temperature    & 0.990 & 0.990 & \textbf{0.995} &  &  &  \\
Rolling percentage     & 0.995 & 0.990 & \textbf{0.995} &  &  &  \\
\cmidrule(lr){1-7}

\multicolumn{7}{l}{\textbf{Recrystallization}} \\
Recrystallization    & 0.980 & \textbf{0.994} & 0.992 & \multirow{3}{*}{0.979} & \multirow{3}{*}{\textbf{0.994}} & \multirow{3}{*}{0.992} \\
Recrystallization  temperature     & 0.971 & \textbf{0.988} & 0.983 &  &  &  \\
Recrystallization time    & 0.971 & \textbf{0.988} & 0.983 &  &  &  \\
\cmidrule(lr){1-7}

\multicolumn{7}{l}{\textbf{Aging}} \\
Aging?            & 0.988 & 0.994 & \textbf{0.995} & \multirow{3}{*}{0.980} & \multirow{3}{*}{0.977} & \multirow{3}{*}{\textbf{0.987}} \\
Aging temperature   & \textbf{0.990} & 0.979 & 0.977 &  &  &  \\
Aging time                 & 0.974 & 0.959 & \textbf{0.986} &  &  &  \\
\midrule

\textbf{Average}                & 0.968 & 0.982 & \textbf{0.986} 
                                & 0.971 & 0.985 & \textbf{0.989} \\
\bottomrule
\end{tabular}
}
\caption{Processing precision of feature-level and tuple-level under different pipelines. Values are rounded to three decimal places; bold numbers indicate the highest precision before rounding.}
\label{tab:process_precision_results}
\end{table}


\begin{table}[htbp]
\centering
\small
\resizebox{\textwidth}{!}{
\begin{tabular}{l|ccc|ccc}
\toprule
 & \multicolumn{3}{c}{Feature level} & \multicolumn{3}{c}{Tuple level} \\
\cmidrule(lr){2-4} \cmidrule(lr){5-7}
 & Single-pass & \makecell{Multi-stage \\ without source} & \makecell{Multi-stage \\ with source} 
 & Single-pass & \makecell{Multi-stage \\ without source} & \makecell{Multi-stage \\ with source} \\
\midrule

\multicolumn{7}{l}{\textbf{Homogenization}} \\
Homogenization?   & \textbf{1.000} & 0.983 & \textbf{1.000} & \multirow{3}{*}{\textbf{1.000}} & \multirow{3}{*}{0.982} & \multirow{3}{*}{\textbf{1.000}} \\
Homogenization temperature        & \textbf{1.000} & 0.986 & \textbf{1.000} &  &  &  \\
Homogenization time     & \textbf{1.000} & 0.986 & \textbf{1.000} &  &  &  \\
\cmidrule(lr){1-7}

\multicolumn{7}{l}{\textbf{Rolling}} \\
Rolling?          & \textbf{1.000} & 0.983 & \textbf{1.000} & \multirow{3}{*}{0.988} & \multirow{3}{*}{0.983} & \multirow{3}{*}{\textbf{1.000}} \\
Rolling temperature               & 0.980 & \textbf{1.000} & \textbf{1.000} &  &  &  \\
Rolling percentage       & 0.980 & \textbf{1.000} & \textbf{1.000} &  &  &  \\
\cmidrule(lr){1-7}

\multicolumn{7}{l}{\textbf{Recrystallization}} \\
Recrystallization    & \textbf{0.991} & 0.983 & 0.987 & \multirow{3}{*}{\textbf{0.988}} & \multirow{3}{*}{0.983} & \multirow{3}{*}{0.984} \\
Recrystallization temperature     & 0.977 & \textbf{0.982} & 0.966 &  &  &  \\
Recrystallization time    & 0.976 & \textbf{0.982} & 0.966 &  &  &  \\
\cmidrule(lr){1-7}

\multicolumn{7}{l}{\textbf{Aging}} \\
Aging?       & 0.997 & 0.980 & \textbf{0.997} & \multirow{3}{*}{0.994} & \multirow{3}{*}{0.980} & \multirow{3}{*}{\textbf{0.995}} \\
Aging temperature        & 0.984 & 0.965 & \textbf{0.995} &  &  &  \\
Aging time                 & 0.984 & 0.964 & 0.982 &  &  &  \\
\midrule

\textbf{Average}                & 0.989 & 0.983 & \textbf{0.991} 
                                & 0.992 & 0.982 & \textbf{0.995} \\
\bottomrule
\end{tabular}
}
\caption{Recall of feature-level and tuple-level under different pipelines for processing.}
\label{tab:process_recall_results}
\end{table}


Similar to the performance observed in the composition extraction, this performance can be attributed to the standardized and structured language. Terminology for processing steps, such as homogenization and rolling, is consistently used across articles with limited lexical variation. Because LLMs rely on statistical associations between words, this kind of formulaic and repetitive phrasing reduces ambiguity and enhances pattern recognition. In addition, processing descriptions tend to follow predictable grammatical structures and are written in imperative or passive voice. Common formats include sequential procedural descriptions, such as: ``After that, the recrystallized samples were further aged at 780\textdegree{}C for 24 h to introduce precipitates.'' They also include conditional statements and parameter descriptions with clearly defined value pairs. These linguistic regularities facilitate automated parsing, as processing steps typically follow a well-structured, logically ordered timeline with explicit quantification.

\subsubsection{Microstructure Features}

Microstructure information extraction presented the biggest challenges in our study due to its complexity (Table \ref{tab:microstructure_precision_results} and Table \ref{tab:microstructure_recall_results}). The single-pass pipeline achieved 0.880 precision and 0.828 recall on microstructure features. Under tuple-level evaluation, the performance further degraded to 0.809 precision and 0.784 recall.

\begin{table}[htbp]
\centering
\small
\resizebox{\textwidth}{!}{
\begin{tabular}{l|ccc|ccc}
\toprule
 & \multicolumn{3}{c}{Feature level} & \multicolumn{3}{c}{Tuple level} \\
\cmidrule(lr){2-4} \cmidrule(lr){5-7}
 & Single-pass & \makecell{Multi-stage \\ without source} & \makecell{Multi-stage \\ with source} 
 & Single-pass & \makecell{Multi-stage \\ without source} & \makecell{Multi-stage \\ with source} \\
\midrule

\multicolumn{7}{l}{\textbf{Matrix}} \\
Matrix type              & 0.923 & 0.977 & \textbf{0.990} & \multirow{2}{*}{0.854} & \multirow{2}{*}{0.844} & \multirow{2}{*}{\textbf{0.900}} \\
Matrix percentage        & \textbf{0.886} & 0.830 & 0.876 &  &  &  \\
\cmidrule(lr){1-7}

\multicolumn{7}{l}{\textbf{Precipitate 1}} \\
Precipitate 1 type       & 0.859 & 0.932 & \textbf{0.980} & \multirow{3}{*}{0.721} & \multirow{3}{*}{0.756} & \multirow{3}{*}{\textbf{0.866}} \\
Precipitate 1 size       & 0.861 & 0.860 & \textbf{0.884} &  &  &  \\
Precipitate 1 percentage & 0.847 & 0.818 & \textbf{0.914} &  &  &  \\
\cmidrule(lr){1-7}

\multicolumn{7}{l}{\textbf{Precipitate 2}} \\
Precipitate 2 type       & 0.839 & 0.881 & \textbf{0.953} & \multirow{3}{*}{0.741} & \multirow{3}{*}{0.672} & \multirow{3}{*}{\textbf{0.841}} \\
Precipitate 2 size       & 0.794 & 0.836 & \textbf{0.879} &  &  &  \\
Precipitate 2 percentage & 0.815 & 0.635 & \textbf{0.878} &  &  &  \\
\cmidrule(lr){1-7}

\multicolumn{7}{l}{\textbf{Precipitate 3}} \\
Precipitate 3 type       & 0.906 & 0.941 & \textbf{0.972} & \multirow{3}{*}{0.919} & \multirow{3}{*}{0.824} & \multirow{3}{*}{\textbf{0.944}} \\
Precipitate 3 size       & 0.950 & 0.895 & \textbf{1.000} &  &  &  \\
Precipitate 3 percentage & \textbf{1.000} & 0.789 & 0.944 &  &  &  \\
\midrule

\textbf{Average}         & 0.880 & 0.854 & \textbf{\textbf{0.934}} 
                         & 0.809 & 0.774 & \textbf{\textbf{0.888}} \\
\bottomrule
\end{tabular}
}
\caption{Precision of feature-level and tuple-level under different pipelines for microstructure.}
\label{tab:microstructure_precision_results}
\end{table}


\begin{table}[htbp]
\centering
\small
\resizebox{\textwidth}{!}{
\begin{tabular}{l|ccc|ccc}
\toprule
 & \multicolumn{3}{c}{Feature level} & \multicolumn{3}{c}{Tuple level} \\
\cmidrule(lr){2-4} \cmidrule(lr){5-7}
 & Single-pass & \makecell{Multi-stage \\ without source} & \makecell{Multi-stage \\ with source} 
 & Single-pass & \makecell{Multi-stage \\ without source} & \makecell{Multi-stage \\ with source} \\
\midrule

\multicolumn{7}{l}{\textbf{Matrix}} \\
Matrix type              & 0.966 & 0.994 & \textbf{0.997} & \multirow{2}{*}{0.885} & \multirow{2}{*}{0.940} & \multirow{2}{*}{\textbf{0.991}} \\
Matrix percentage        & 0.886 & 0.919 & \textbf{0.989} &  &  &  \\
\cmidrule(lr){1-7}

\multicolumn{7}{l}{\textbf{Precipitate 1}} \\
Precipitate 1 type       & 0.926 & 0.977 & \textbf{0.997} & \multirow{3}{*}{0.836} & \multirow{3}{*}{0.928} & \multirow{3}{*}{\textbf{0.977}} \\
Precipitate 1 size       & 0.892 & 0.968 & \textbf{0.986} &  &  &  \\
Precipitate 1 percentage & 0.853 & 0.935 & \textbf{0.983} &  &  &  \\
\cmidrule(lr){1-7}

\multicolumn{7}{l}{\textbf{Precipitate 2}} \\
Precipitate 2 type       & 0.803 & 0.912 & \textbf{0.968} & \multirow{3}{*}{0.692} & \multirow{3}{*}{0.813} & \multirow{3}{*}{\textbf{0.946}} \\
Precipitate 2 size       & 0.685 & 0.864 & \textbf{0.906} &  &  &  \\
Precipitate 2 percentage & 0.815 & 0.870 & \textbf{0.985} &  &  &  \\
\cmidrule(lr){1-7}

\multicolumn{7}{l}{\textbf{Precipitate 3}} \\
Precipitate 3 type       & 0.690 & 0.865 & \textbf{0.946} & \multirow{3}{*}{0.723} & \multirow{3}{*}{0.848} & \multirow{3}{*}{\textbf{0.944}} \\
Precipitate 3 size       & 0.760 & 0.895 & \textbf{0.950} &  &  &  \\
Precipitate 3 percentage & 0.826 & \textbf{1.000} & 0.944 &  &  &  \\
\midrule

\textbf{Average}         & 0.828 & 0.927 & \textbf{0.968} 
                         & 0.784 & 0.882 & \textbf{0.965} \\
\bottomrule
\end{tabular}
}
\caption{Recall of feature-level and tuple-level under different pipelines for microstructure.}
\label{tab:microstructure_recall_results}
\end{table}



This performance gap compared to other feature categories led us to investigate the underlying causes:

\begin{itemize}
\item Cognitive load and task complexity issues: Single-pass pipeline requires LLM to complete multiple complex tasks in one processing pass: identifying all materials, extracting 47 features, understanding interdependencies between features, and correctly associating scattered information. As a result, the LLMs experienced cognitive overload.
\item Inherent hierarchical structure: Microstructure identification often depends on composition and processing conditions. While single-pass pipeline ``sees'' all information, it may not have sufficiently ``understood'' compositional and processing information when collecting microstructure information.
\item Attention mechanism limitations: Information scattered across different sections of papers. Single-pass requires the model to simultaneously track multidimensional information for multiple materials across thousands of words.
\end{itemize}

Multi-stage pipeline overcomes these problems by first extracting composition and processing, then microstructure, followed by properties, and finally performing a validation of all extracted features against the original text. During the staged extraction process, we provide both the already-extracted features and the original text. This extraction sequence follows the writing order of most material science articles.

This contextual enrichment mechanism slightly improved performance, with feature-level evaluation results showing 0.854 precision and 0.927 recall. Compared to the single-pass pipeline, the multi-stage pipeline without source tracking achieved 2.6\% lower accuracy and 10.1\% higher recall. Specifically, matrix type and each precipitate type showed an average precision improvement of 5.1\%, while matrix percentage, precipitate size, and precipitate percentage showed minimal precision improvements or even significant decreases; matrix type and precipitate type showed an average recall improvement of 9.1\%, while matrix, precipitate type, size, and percentage displayed an average recall improvement of 10.5\%. Tuple-level evaluation results showed 0.774 precision and 0.882 recall. Compared to the single-pass pipeline, the multi-stage pipeline without source tracking exhibited similar magnitudes of change in overall accuracy, recall, and the accuracy and recall of individual components. 

The multi-stage approach reduced the complexity of each task and focused on specific information types at each stage, significantly enhancing the model's recall performance. Although significant improvements were achieved in recall, accuracy decreased by approximately 2\%. Specifically, matrix type and precipitate type showed improved precision and recall with good post-improvement performance. However, matrix percentage, precipitate percentage, and size exhibited minimal precision improvements. This pattern reveals why recall increased while precision slightly decreased. The multi-stage pipeline became more likely to attempt extraction of unreported values,  successfully capturing many previously missed true positives (reducing false negatives), but the model also predicted more incorrect ``has values,'' increasing false positives and slightly decreasing precision.

To address these precision issues while maintaining the improved recall, we implemented source tracking in our multi-stage pipeline. Source tracking enhances attention by improving signal-to-noise ratio through better attention patterns \cite{vaswani2017attention,wang2024trace} and prevents cascade errors while enabling precise error localization during validation \cite{caselli2015s}.  This pipeline achieved significant performance improvements in microstructure extraction. In feature-level evaluation, it achieved 0.934 precision and 0.968 recall, improving by 8.0\% and 4.1\% respectively compared22 to the multi-stage pipeline without source tracking, and by 5.4\% and 14.1\% compared to single-pass extraction. In tuple-level evaluation, performance reached 0.888 precision and 0.965 recall, representing improvements of 11.4\% and 8.3\% over the non-source-tracking multi-stage approach, and 7.9\% and 18.1\% over single-pass extraction.

A large portion of the errors is still caused by the model's inference attempts. When the model performed reasoning, it sometimes marked outputs as ``estimated,'' or included intermediate calculation steps in its reasoning process. Following strict evaluation criteria, we counted these incorrectly inferred values as false positives even when explicitly marked as estimates. This is because our basic facts define positive examples as values that are clearly present in the original text or correct values obtained through reasoning. However, the multi-stage pipeline with source tracking already showed a considerable reduction in erroneous or unnecessary inference compared to the other two pipelines.

The multi-stage pipeline with source tracking thus represents an optimal balance. It maintains the high recall achieved through staged extraction while dramatically improving precision through source anchoring. This approach successfully extracted 96.5\% of all true microstructure values in tuple-level evaluation. Even when unnecessary or erroneous inference leads to false positives, these can be filtered out by discarding values marked as inferred.

\subsubsection{Property Features}

Property extraction achieved strong performance across all pipelines (Fig. \ref{tab:properties_results}). The multi-stage pipeline with source tracking reached 0.941 precision and 0.991 recall. Compared to single-pass extraction, this represents a shift: recall increased significantly by 9.5\% (from 0.896 to 0.991) while precision decreased by 3.7\% (from 0.941 to 0.978).

\begin{table}[htbp]
\centering
\small
\resizebox{\textwidth}{!}{
\begin{tabular}{l|ccc|ccc}
\toprule
 & \multicolumn{3}{c}{Precision} & \multicolumn{3}{c}{Recall} \\
\cmidrule(lr){2-4} \cmidrule(lr){5-7}
 & Single-pass & \makecell{Multi-stage \\ without source} & \makecell{Multi-stage \\ with source} 
 & Single-pass & \makecell{Multi-stage \\ without source} & \makecell{Multi-stage \\ with source} \\
\midrule
Ultimate Tensile Strength 
 & 0.976 & 0.967 & \textbf{0.981} 
 & 0.976 & 0.955 & \textbf{0.994} \\
Ultimate Compressive Strength 
 & \textbf{1.000} & \textbf{1.000} & \textbf{1.000} 
 & \textbf{1.000} & 0.957 & \textbf{1.000} \\
Tensile Yield Strength 
 & 0.950 & 0.947 & \textbf{0.956} 
 & 0.972 & 0.976 & \textbf{0.994} \\
Compressive Yield Strength 
 & \textbf{1.000} & 0.972 & 0.927 
 & 0.929 & 0.921 & \textbf{1.000} \\
Hardness 
 & \textbf{1.000} & 0.951 & 0.951 
 & 0.891 & \textbf{0.942} & 0.924 \\
Tensile Ductility 
 & 0.958 & 0.956 & \textbf{0.982} 
 & 0.941 & 0.956 & \textbf{1.000} \\
Compressive Ductility 
 & \textbf{1.000} & \textbf{1.000} & \textbf{1.000} 
 & \textbf{1.000} & \textbf{1.000} & \textbf{1.000} \\
\makecell{NRT-Cryo Strength Heat-treated Temperature}
 & \textbf{0.967} & 0.955 & 0.895 
 & 0.569 & 0.857 & \textbf{1.000} \\
NRT-Cryo Heat-treated Strength 
 & \textbf{1.000} & 0.955 & 0.833 
 & 0.862 & 0.875 & \textbf{1.000} \\
NRT-Cryo Heat-treated Ductility 
 & 0.933 & \textbf{1.000} & 0.882 
 & 0.824 & \textbf{1.000} & \textbf{1.000} \\
\midrule
\textbf{Average} 
 & \textbf{0.978} & 0.970 & 0.941 
 & 0.896 & 0.944 & \textbf{0.991} \\
\bottomrule
\end{tabular}
}
\caption{Precision and recall of different pipelines for property.}
\label{tab:properties_results}
\end{table}


Analysis of the 40 property extraction errors from the multi-stage pipeline with source tracking reveals three primary failure modes. 

Confusion errors, comprising 62.5\% of total errors, predominantly involve attributing properties from one material to another. Three distinct patterns emerged from our analysis. First, academic terminology creates interpretation challenges. When encountering ``The hardness value increases from the solutionized condition of 330 $\pm$ 8 HV to a peak value of 380 $\pm$ 5 HV after 50 h of aging,'' the model failed to recognize ``solutionized condition'' as the initial state (0 h), instead treating the 330 HV value as belonging to an intermediate aging condition. Second, incomplete property reporting across multiple samples led to mis-attribution. When papers focus on specific phenomena such as long-term stability, they may report complete property sets only for key conditions while providing partial data for others. The model, attempting to maintain completeness, occasionally misattributed properties from well-characterized samples to incompletely characterized ones. Third, contextual emphasis created confusion. When a paper stated ``large ductility over 22\% and good phenomenally strain-hardening capability over 400 MPa can be well retained even at a high yield strength approaching 1.0 GPa,'' the model incorrectly extracted 1.0 GPa as the material's exact yield strength, failing to recognize that the emphasis was on the 22\% ductility retention at high strength levels, with the actual yield strength reported elsewhere.

Extra inference errors, comprising 12.5\% of all failures, resulted from the LLM inappropriately applying theoretical models rather than limiting itself to data re-expression. A telling example occurred when the text stated ``the base strength (i.e., unstrengthened matrix) is 165 MPa and the L1$_2$ particle shear strengthening contribution is 244.3 MPa.'' LLM incorrectly derived a total strength of 409 MPa (165 + 244) by applying a linear addition strengthening model, treating this theoretical prediction as an experimental value. While reasonable inferences such as calculating atomic percentages from chemical formulae (e.g., Fe$_2$Al$_5$ $\rightarrow$ Fe: 28.6 at\%, Al: 71.4 at\%) represent legitimate re-expression of experimental data, this case involved unauthorized theoretical modeling that crossed the boundary from data transformation into model prediction. The challenge lies in controlling LLM behavior to distinguish between appropriate format conversions and inappropriate theoretical calculations. 

Omission errors constitute 25\% of the total and primarily result from indirect property reporting. For example, when papers use expressions like ``1200 MPa higher than the ST condition'' without explicitly stating the baseline value, the model fails to extract this information, unlike when values are directly stated, such as ``\~274 MPa'' for yield strength. Models perform unnecessary computations in inference errors and fail to perform necessary computations in omissions. This means that the model lacks the domain knowledge to understand when such calculations are physically meaningful.

\section{Discussion}\label{sec3}

\subsection{Multi-stage Pipeline}

The multi-stage architecture's advantages arise from the hierarchical dependencies inherent in materials science. Composition and processing collectively affect microstructures, which in turn play a significant role in governing properties. Our four-stage design mirrors typical materials science paper structure and enables progressive construction of complete material information graphs. The systematic extraction of 47 features across the entire composition-process-microstructure-property chain represents unprecedented comprehensiveness. This approach preserves not just individual features but their critical interconnections.

Building on this architectural foundation, we next introduce the role of source tracking in ensuring reliability. Source tracking introduces a cascade error correction mechanism essential for maintaining extraction fidelity. By preserving the original text context, the source text serves as an anchor and avoids reference loss. When encountering ambiguous references like ``the FSA specimen,'' the associated source text maintains proper material identification across extraction stages. During final validation, the system cross-references all extracted information against original sources, enabling identification and correction of propagated errors. This retrospective validation capability explains our achievement of zero false positive materials while maintaining exceptional precision. 

\subsection{Performance and Advantages}

The integration of source tracking with multi-stage extraction represents a fundamental advance in automated scientific information extraction. The reduction in missed materials from 43 (10.8\%) to 13 (3.3\%) indicates the positive impact of source tracking. The impact is most pronounced in microstructure extraction, where tuple-level evaluation showed precision improving from 0.809 to 0.888 and recall from 0.784 to 0.965. Composition and processing extraction achieved over 98\% accuracy across all pipelines. This high performance highlights LLMs' proficiency with standardized scientific language. The success can be attributed to the formulaic nature of chemical notation and processing descriptions, which follow predictable patterns that align well with LLM training. Source tracking's effectiveness further shows sophisticated contextual understanding --- the model correctly resolves material references, interprets abbreviations, and maintains entity consistency when provided with anchoring text.

Our evaluation reveals both capabilities and limitations of current LLMs in materials science. For example, when processing microstructure type encoding, the system needed to recognize diverse expressions such as \(\sigma\) phase, Eutectic \(\gamma + \gamma'\) phase, \(\varepsilon\)-phase, and \(\gamma'\) precipitates, then determine whether these correspond to our predefined encoding system (L12, \(\gamma''\), B2, etc.). This task requires not merely surface text understanding but mastery of crystallographic nomenclature, microstructure knowledge, and materials science terminology. The model's success in navigating these complexities while maintaining high accuracy reflects deep domain knowledge.

\subsection{Limitations and Challenges}

However, the model exhibited tendencies toward inappropriate inference when it handled incomplete information. When the text stated ``precipitate A comprises 20\% volume fraction,'' the model often assumed the matrix occupied 80\% and ignored potentially unreported phases. This completion bias reflects LLMs' training on complete datasets but conflicts with scientific reporting where ``unreported'' differs fundamentally from ``non-existent.''

Domain-specific reasoning gaps manifested across multiple feature categories, most prominently in microstructure and property extraction. For example, when encountering aging data at different temperatures, the system sometimes performed unsupported extrapolation. For instance, given ``L12 precipitates are \textasciitilde5 nm at 550\degree C/150h and \textasciitilde75 nm at 700\degree C/50h,'' the model might infer intermediate sizes at 800\degree C/20h despite lacking empirical support for such interpolation. The system also struggled with mixed phase classification: when literature described ``mixed BCC/B2 phase,'' it needed to determine whether this represented two separate precipitates or a single mixed-structure precipitate --- a decision requiring crystallographic expertise the model lacked.

Further challenges arose in hierarchical information aggregation. When papers reported microstructure information for different regions (``Region A: 15\% \(\sigma\) phase'' and ``Region B: 8\% L12 phase''), the system had to decide whether to report regional data separately or attempt calculation of bulk properties. Errors occurred when the model inappropriately generalized regional information to represent the entire material or failed to properly weigh regional contributions. The conversion between qualitative and quantitative descriptions posed additional difficulties. The model struggled with semi-quantitative expressions (``less than 5\%,'' ``approximately mid-to-low 300 MPa'') and purely qualitative terms (``very fine,'' ``dominant phase''), sometimes inappropriately assigning specific numerical values to inherently imprecise descriptions.

Similar reasoning deficits appeared in property extraction, where 12.5\% of errors involved inappropriate theoretical modeling rather than limiting extraction to experimental data. As discussed earlier, the extraction system applied theoretical models to derive predicted values that were never experimentally measured. These parallel failures across microstructure and properties extraction reveal a fundamental limitation: while LLMs excel at pattern recognition and information retrieval, they struggle to distinguish between experimental measurements and theoretical predictions in scientific literature. These findings point to a future research direction: developing better strategies to control LLM behavior and ensure strict adherence to experimental data extraction without unauthorized theoretical modeling.

\subsection{Applications and Future Directions}

Our multi-stage source-tracking approach offers several advantages for material information extraction. Compared with manual extraction, the system significantly reduces processing time while maintaining a high F1 score and accuracy, making large-scale literature mining practically feasible. Our 47-feature framework captures a relatively complete material feature space in a unified structured format, including composition (14 features), processing (12 features), structure (11 features), and property (10 features). The approach delivers a low material miss rate and zero false positives. This low error rate ensures that subsequent machine learning models can be trained without suffering from performance degradation caused by noisy datasets. Despite being demonstrated on alloys, the pipeline generalizes broadly across material types, as the composition-process-microstructure-property relationships and modular feature design enable adaptation through feature modification while maintaining the core framework. In addition, the extracted structured data enables knowledge graph construction in order to support cross-system knowledge transfer and hypothesis generation.

Current limitations center on several areas requiring future development. The system processes only text and tables, missing information embedded in images. Future systems will integrate multimodal capabilities for comprehensive extraction. Additionally, our current pipeline consumes significant computational resources, with an average of approximately 21,500 tokens per paper (including both input and output), as each stage requires inputting the complete original text plus all previously extracted information to the LLM. Future work should develop more cost-effective solutions, potentially through selective context windowing or incremental processing strategies.

\section{Methods}\label{sec4}
\subsection{Dataset Construction}

100 journal articles related to precipitate-containing multi-principal element alloys were collected for testing the information extraction pipelines. Two materials science experts annotated the dataset, extracting a comprehensive set of 47 features across four distinct categories from 396 materials: composition (14 features), processing parameters (12 features), microstructure (11 features), and material properties (10 features).

\subsection{Pre-processing Setup}

PDF documents were processed using PyMuPDF (fitz) for plain text extraction. Although this approach does not preserve table formatting, the numerical data and text content from tables are still extracted and remain interpretable by the LLM. Figures and images were excluded as the o3-mini API only accepts text-based inputs. The extracted content maintained its original formatting as produced by PyMuPDF without additional standardization or character normalization. All extractions utilized the OpenAI o3-mini API with reasoning\_effort parameter configured as ``high''. Output was structured in JSON format to ensure consistent parsing and validation.

\subsection{Four-Stage Extraction Pipeline}

Our extraction pipeline employs a systematic four-stage approach designed to leverage the hierarchical nature of material information. Each extraction stage employs a dual structure: predefined fields for standard information categories and corresponding ``Additional'' fields (Other\_Composition, Additional\_Processing, Additional\_Microstructure, Additional\_Properties) to ensure no reported information is lost due to rigid categorization.

Stage 1 performs initial material identification and extraction across the entire document. In this stage, we extract all materials mentioned in the article along with their chemical compositions in atomic percentages and processing parameters. Processing information includes homogenization conditions with temperature and time, rolling parameters encompassing temperature and thickness reduction percentage, recrystallization conditions with temperature and time, and aging treatments with their corresponding temperature and duration. Each material is assigned to a unique identifier, and all extracted values are accompanied by their source text from the original document. The extracted information is then converted to JSON format for structured data processing.

Following Stage 1, we implement an iterative extraction process for each identified material. For each material, Stage 2 first extracts microstructure information. This stage receives the specific material's composition and processing data from Stage 1, along with source texts, extraction guidelines, and the full article text. The extraction targets matrix phase type and volume fraction, along with precipitate phases through structured fields capturing their types, sizes, and volume fractions. Corresponding source texts are recorded for each extracted phase feature, and the microstructure data is converted to JSON format.

Stage 3 immediately follows Stage 2 for the same material, extracting its mechanical properties. Building upon the accumulated composition, processing, and microstructure information for this specific material, Stage 3 captures room temperature mechanical properties, including ultimate tensile and compressive strength, yield strength, hardness, and ductility values. Additionally, non-ambient temperature properties are captured, specifically strength and ductility measurements at elevated or cryogenic temperatures. All property values are linked to their source texts, completing the information extraction for this individual material, with outputs converted to JSON format.

This Stage 2--3 sequence repeats for each material identified in Stage 1, ensuring that microstructure and property information are extracted with the full context of each material's specific composition and processing history. The iterative approach maintains material-specific context throughout the extraction process, reducing cross-material contamination and improving attribution accuracy.

Stage 4 performs confirmation only after all materials have been fully processed through Stages 2 and 3. Using the complete material database containing all extracted materials with their composition, processing, microstructure, and property information, along with all source texts and the full article text, the API conducts this final confirmation stage to verify extraction accuracy and consistency across the entire dataset. The confirmation process checks phase volume percentages, ensures properties align with their corresponding processing conditions, verifies material uniqueness, and performs overall quality control.

\subsection{Prompt Engineering Strategies}

Three core strategies guide our prompt design to ensure accurate and complete extraction.

The first strategy involves the definition of a unique material. Each unique combination of composition and processing conditions constitutes a separate material entry. This prevents conflation of materials with identical compositions but different processing histories.

The second strategy implements controlled flexibility through structured alternative fields. The system includes provisions for ``Other\_Composition'', ``Additional\_Processing'', ``Additional\_Microstructure'', and ``Additional\_Properties'' to ensure comprehensive information capture beyond predefined categories. This design recognizes that materials science literature often reports information that doesn't conform to standard classifications. For composition, while we track 14 primary elements, the ``Other\_Composition'' field captures trace elements. For microstructure information, we provide three structured fields for primary precipitates with detailed attributes (type, size, volume fraction), but the ``Additional\_Microstructure'' field ensures that papers reporting four, five, or more precipitate types can be fully captured without forcing data into inappropriate categories. Similarly, ``Additional\_Processing'' accommodates novel or non-standard processing steps beyond homogenization, rolling, recrystallization, and aging. ``Additional\_Properties'' captures specialized mechanical or physical properties beyond our core set of strength, hardness, and ductility measurements. This flexible architecture prevents information loss while maintaining structured data organization, achieving both completeness and consistency in extraction.

The third strategy emphasizes comprehensive source tracking. Each extraction stage requires preservation of the source text, which is passed forward to subsequent stages. This creates a cumulative context that reduces ambiguity in material references, enables verification of extracted values, and maintains traceability throughout the extraction pipeline.

\subsection{Reasoning Configuration}

We configured the extraction system with \texttt{reasoning\_effort=``high''} without restricting the model's reasoning capabilities in prompts. This setting enables the model to invest additional computational resources for deep reasoning, construct complete logical chains, carefully analyze complex causal relationships, and provide well-founded inferences rather than arbitrary guesses when facing ambiguous information.

This approach proves particularly valuable in the materials science literature where certain calculations and inferences are routine. For example, when a paper provides only the chemical formula Fe$_2$Al$_5$ without explicit atomic percentages, the model accurately calculates Fe: 28.6~at\%, Al: 71.4~at\%. Similarly, when papers describe materials containing a matrix phase and two precipitate phases with only partial volume fraction information, such as matrix at 85\% and one precipitate at 10\%, the model correctly infers that the second precipitate occupies 5\% of the volume.

\subsection{Evaluation Metrics}

Building on standard classification metrics, our evaluation addresses the challenges of materials information extraction through a two-tiered approach. First, we evaluate performance only for correctly identified materials to provide insight into extraction quality. Then we penalize the system for any missed materials (counting all 47 features as false negatives) or erroneously extracted materials (counting all 47 features as false positives). For processing steps and microstructure information, we implement the two complementary calculation approaches described in Section 2.3: single-feature evaluation and tuple-based evaluation.

To evaluate extraction performance, we use standard classification metrics based on confusion matrix elements. For each evaluation task, we define True Positives (TP) as correctly extracted features or materials that match the ground truth, False Positives (FP) as incorrectly extracted features or materials not present in the ground truth, True Negatives (TN) as correctly identified absent features or materials, and False Negatives (FN) as ground truth features or materials that were missed by the extraction system.

Based on these fundamental counts, the evaluation metrics are calculated as follows:

\begin{itemize}
\item
Precision measures the proportion of extracted items that are correct:
\begin{equation*}
\text{Precision} = \frac{\mathrm{TP}}{\mathrm{TP} + \mathrm{FP}} 
= \frac{\text{Number of correctly predicted positives}}{\text{Total number of extractions}}
\end{equation*}
\item 
Recall measures the proportion of ground truth items that are successfully extracted:
\begin{equation*}
\text{Recall} = \frac{\mathrm{TP}}{\mathrm{TP} + \mathrm{FN}} 
= \frac{\text{Number of correct extractions}}{\text{Total number in ground truth}}
\end{equation*}
\item
F1 Score provides the harmonic mean of precision and recall:
\begin{equation*}
\text{F1} = \frac{2 \times \text{Precision} \times \text{Recall}}{\text{Precision} + \text{Recall}} 
= \frac{2\mathrm{TP}}{2\mathrm{TP} + \mathrm{FP} + \mathrm{FN}}
\end{equation*}
\item
Accuracy measures the overall correctness across all predictions:
\begin{equation*}
\text{Accuracy} = \frac{\mathrm{TP} + \mathrm{TN}}{\mathrm{TP} + \mathrm{TN} + \mathrm{FP} + \mathrm{FN}} 
= \frac{\text{Number of correctly predicted instances}}{\text{Total number of predictions}}
\end{equation*}
\end{itemize}

\bibliography{sn-article}

\backmatter

\section*{Data and Code availability}
All code and data will be made available upon request. Research articles used for evaluating the pipeline in this work were obtained through institutional licensing agreements between the University of Oklahoma and major academic publishers, including Elsevier, Wiley, American Chemical Society (ACS), Springer Nature, Taylor \& Francis, among others. These agreements ensure legitimate access to the scientific literature required for comprehensive evaluation of our pipelines.

\section*{Acknowledgements}

Some of the computing for this project was performed at the OU Supercomputing Center for Education \& Research (OSCER). X.W.\ and A.R.\ acknowledge support from the Graduate Student Seed Funding Program of the Data Institute for Societal Challenges at the University of Oklahoma. H.W.\ acknowledges the financial support by the U.S.\ Nuclear Regulatory Commission Faculty Development Program (award number NRC 31310018M0044). S.X.\ was supported by a grant from the Research Council of the University of Oklahoma (OU) Norman Campus. 

\section*{Author contribution}
S.X.\ and K.L.\ conceived the study. X.W.\ performed pipelines development, testing, and analysis. A.R., M.L., H.W., S.X., and X.W.\ completed dataset creation and validation. K.L.\ supervised the research and designed the task evaluation metrics. X.W.\ wrote the manuscript, and all authors contributed to manuscript revision and editing.

\section*{Competing interests}
The authors declare no competing interests.

\end{document}